\documentclass{nle}
\usepackage[utf8]{inputenc}
\usepackage{graphicx}
\usepackage{xcolor}
\usepackage{url}
\usepackage{amsmath}
\usepackage[capitalize,nameinlink]{cleveref}
\usepackage{wrapfig}

\usepackage[normalem]{ulem}

\makeatletter
\let\O@argtabularcr\@argtabularcr
\def\O@xtabularcr{\@ifnextchar[\O@argtabularcr{\ifnum 0=`{\fi}\cr}}
\let\O@tabacol\@tabacol
\let\O@tabclassiv\@tabclassiv
\let\O@tabclassz\@tabclassz
\let\O@tabarray\@tabarray
\def\author@tabular{\authorsize\def\@halignto{}\@authortable}
\let\endauthor@tabular=\endtabular
\def\author@tabcrone{{\ifnum0=`}\fi\O@xtabularcr\affilsize\itshape
 \let\\=\author@tabcrtwo\ignorespaces}
\def\author@tabcrtwo{{\ifnum0=`}\fi\O@xtabularcr[-3\p@]\affilsize\itshape
 \let\\=\author@tabcrtwo\ignorespaces}
\def\@authortable{\leavevmode \hbox \bgroup $\let\@acol\O@tabacol
 \let\@classz\O@tabclassz \let\@classiv\O@tabclassiv
 \let\\=\author@tabcrone \ignorespaces \O@tabarray}
\makeatother

\usepackage{colortbl}

\usepackage{collcell}
\usepackage{hhline}
\usepackage{pgf}
\usepackage{multirow}

\def\colorModel{hsb} 

\newcolumntype{E}{>{\collectcell\ColCellSNLI}c<{\endcollectcell}}  
\newcommand\ColCellSNLI[1]{
  \pgfmathparse{#1<9824/4?1:0}  
    \ifnum\pgfmathresult=0\relax\color{white}\fi
  \pgfmathsetmacro\compA{0}       
  \pgfmathsetmacro\compB{#1/(9824/2)} 
  \pgfmathsetmacro\compC{1}       
  \edef\x{\noexpand\centering\noexpand\cellcolor[\colorModel]{\compA,\compB,\compC}}\x #1
  }
\newcolumntype{F}{>{\collectcell\ColCellMNLI}c<{\endcollectcell}}  
\newcommand\ColCellMNLI[1]{
  \pgfmathparse{#1<9832/4?1:0}  
    \ifnum\pgfmathresult=0\relax\color{white}\fi
  \pgfmathsetmacro\compA{0}       
  \pgfmathsetmacro\compB{#1/(9832/2)} 
  \pgfmathsetmacro\compC{1}       
  \edef\x{\noexpand\centering\noexpand\cellcolor[\colorModel]{\compA,\compB,\compC}}\x #1
  }
\newcolumntype{G}{>{\collectcell\ColCellSciNLI}c<{\endcollectcell}}  
\newcommand\ColCellSciNLI[1]{
  \pgfmathparse{#1<2400/4?1:0}  
    \ifnum\pgfmathresult=0\relax\color{white}\fi
  \pgfmathsetmacro\compA{0}       
  \pgfmathsetmacro\compB{#1/(2400/2)} 
  \pgfmathsetmacro\compC{1}       
  \edef\x{\noexpand\centering\noexpand\cellcolor[\colorModel]{\compA,\compB,\compC}}\x #1
  }

\title[Sentence Embeddings in NLI with Iterative Refinement Encoders]
      {Sentence Embeddings in NLI with Iterative Refinement Encoders}

\author[Talman, Yli-Jyr\"a and Tiedemann]
       {Aarne Talman, Anssi Yli-Jyr\"a and  J\"org Tiedemann\\
       Department of Digital Humanities\\University of Helsinki \\
{\tt \{aarne.talman, anssi.yli-jyra, jorg.tiedemann\}@helsinki.fi}}

\received{}

\pagerange{\pageref{firstpage}--\pageref{lastpage}}
\pubyear{2019}

\usepackage{nle-chicago-patch}
\bibpunct{(}{)}{,}{a}{}{;}

\begin{document}

\label{firstpage}
\maketitle

\begin{abstract}

Sentence-level representations are necessary for various NLP tasks. Recurrent neural networks have proven to be very effective in learning distributed representations and can be trained efficiently on natural language inference tasks. We build on top of one such model and propose a hierarchy of BiLSTM and max pooling layers that implements an iterative refinement strategy and yields state of the art results on the SciTail dataset as well as strong results for SNLI and MultiNLI. We can show that the sentence embeddings learned in this way can be utilized in a wide variety of transfer learning tasks, outperforming InferSent on 7 out of 10 and SkipThought on 8 out of 9 SentEval sentence embedding evaluation tasks. Furthermore, our model beats the InferSent model in 8 out of 10 recently published SentEval probing tasks designed to evaluate sentence embeddings' ability to capture some of the important linguistic properties of sentences.
\end{abstract}

\section{Introduction}

Neural networks have been shown to provide a powerful tool for building representations of natural languages on multiple levels of linguistic abstraction.  Perhaps the most widely used representations in natural language processing are word embeddings \citep{Mikolov:2013,pennington2014glove}.  Recently there has been a growing interest in models for sentence-level representations using a range of different neural network architectures.  Such {\em sentence embeddings} have been generated using unsupervised learning approaches \citep{KirosZSZTUF15,HillCK16}, and supervised learning \citep{BowmanGRGMP16,infersent}.

Supervision typically comes in the form of an underlying semantic task with labeled data to train the model. The most prominent task for that purpose is natural language inference (NLI) that tries to model the inferential relationship between two or more given sentences. In particular, given two sentences - the premise $p$ and the hypothesis $h$ - the task is to determine whether $h$ is entailed by $p$, whether the sentences are in contradiction with each other or whether there is no inferential relationship between the sentences (neutral). There are two main neural approaches to NLI. {\em Sentence encoding-based models} focus on building separate embeddings for the premises and the hypothesis and then combine those using a classifier \citep{snli,BowmanGRGMP16,infersent}. Other approaches do not treat the two sentences separately but utilize e.g. cross-sentence attention \citep{cafe,Chen}.

With the goal of obtaining general-purpose sentence representations in mind, we opt for the sentence encoding approach. Motivated by the success of the InferSent architecture \citep{infersent} we extend their architecture with a hierarchy-like structure of bidirectional LSTM (BiLSTM) layers with max pooling. All in all, our model improves the previous state of the art for SciTail \citep{scitail} and achieves strong results for the SNLI and Multi-Genre Natural Language Inference corpus (MultiNLI; \citealp{multinli}).

In order to demonstrate the semantic abstractions achieved by our approach, we also apply our model to a number of transfer learning tasks using the SentEval testing library \citep{infersent}, and show that it outperforms the InferSent model on 7 out of 10 and SkipThought \citep{KirosZSZTUF15} on 8 out of 9 tasks, comparing to the scores reported by \cite{infersent}. Moreover, our model outperforms the InferSent model in 8 out of 10 recently published SentEval probing tasks designed to evaluate sentence embeddings' ability to capture some of the important linguistic properties of sentences \citep{probing}.  This highlights the generalization capability of the proposed model, confirming that its architecture is able to learn sentence representations with strong performance across a wide variety of different NLP tasks.

\section{Related Work}
There is a wide variety of approaches to sentence-level representations that can be used in natural language inference. \cite{snli} and \cite{BowmanGRGMP16} explore RNN and LSTM architectures, \cite{MouConv} convolutional neural networks and \cite{VendrovKFU15} GRUs, to name a few. The basic idea behind these approaches is to encode the premise and hypothesis sentences separately and then combine those using a neural network classifier.

\cite{infersent} explore multiple different sentence embedding architectures ranging from LSTM, BiLSTM and intra-attention to convolution neural networks and the performance of these architectures on NLI tasks. They show that, out of these models, BiLSTM with max pooling achieves the strongest results not only in NLI but also in many other NLP tasks requiring sentence level meaning representations. They also show that their model trained on NLI data achieves strong performance on various transfer learning tasks.

Although sentence embedding approaches have proven their effectiveness in NLI, there are multiple studies showing that treating the hypothesis and premise sentences together and focusing on the relationship between those sentences yields better results \citep{cafe,Chen}.
These methods are focused on the inference relations rather than the internal semantics of the sentences.
Therefore, they do not offer similar insights about the sentence level semantics, as individual sentence embeddings do, and they cannot  straightforwardly be used outside of the NLI context.

\section{Model Architecture}
\label{sec:architecture}

\begin{figure}[h]
\begin{center}
  \includegraphics[width=0.5\linewidth]{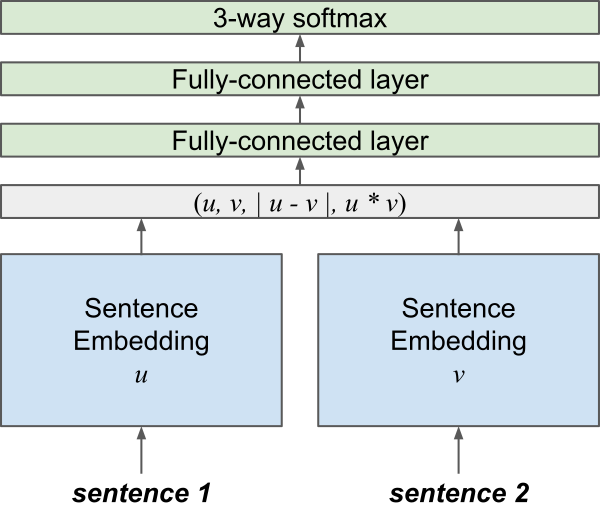}
\end{center}
  \caption{Overall NLI Architecture}
  \label{fig:nli}
\end{figure}

Our proposed architecture follows a sentence embedding-based approach for NLI introduced by \cite{snli}. The model illustrated in Figure \ref{fig:nli} contains sentence embeddings for the two input sentences, where the output of the sentence embeddings are combined using a heuristic introduced by \cite{MouConv}, putting together the concatenation $(u,v)$, absolute element-wise difference $|u-v|$, and element-wise product $u*v$. The combined vector is then passed on to a 3-layered multi-layer perceptron (MLP) with a 3-way softmax classifier. The first two layers of the MLP both utilize dropout and a ReLU activation function.

We use a variant of ReLU called Leaky ReLU \citep{Maas2013RectifierNI}, defined by:
$$LeakyReLU(x) = \max(0, x) + y * \min(0, x)$$
where we set $y = 0.01$ as the negative slope for $x < 0$.
This prevents the gradient from dying when $x < 0$.

For the sentence representations we first embed the individual words with pre-trained word embeddings. The sequence of the embedded words is then passed on to the sentence encoder which utilizes BiLSTM with max pooling. Given a sequence $T$ of words $(w_1\ldots,w_T)$, the output of the bi-directional LSTM is
a set of vectors $(h_1,\ldots,h_T)$, where each $h_t\in(h_1,\ldots,h_T)$ is the concatenation
$$h_t = [\overrightarrow{h}_t,\overleftarrow{h}_t]$$
of a forward and backward LSTMs
$$\overrightarrow{h}_t = \overrightarrow{LSTM}_t(w_1,\ldots,w_t)$$
$$\overleftarrow{h}_t = \overleftarrow{LSTM}_t(w_T,\ldots,w_t).$$
The max pooling layer produces a vector of the same dimensionality as $h_t$, returning, for each dimension, its maximum value over
the hidden units $(h_1,\ldots,h_T)$.

\begin{figure}[t]
  \includegraphics[width=0.9\linewidth]{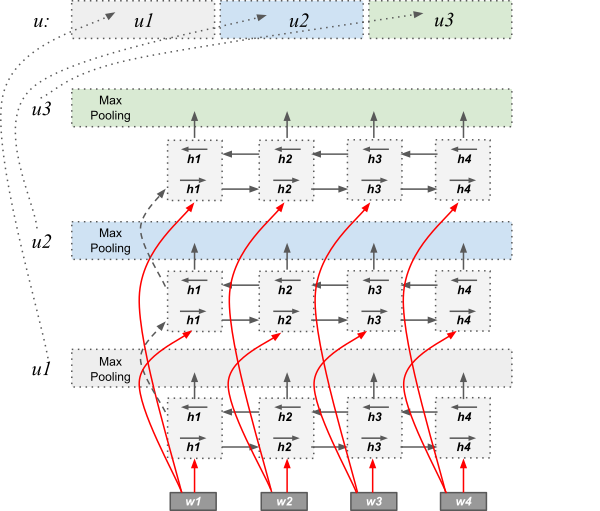}
  \caption{Architecture of the HBMP sentence encoder (where $T=4$).}
  \label{fig:architecture}
\end{figure}

Motivated by the strong results of the BiLSTM max pooling network by \cite{infersent}, we experimented with combining BiLSTM max pooling networks in a hierarchy-like structure.\footnote{\cite{infersent} explore a similar architecture using convolutional neural networks, called Hierarchical ConvNet.} To improve the BiLSTM layers' ability to remember the input words, we let each layer of the network re-read the input embeddings instead of stacking the layers in a strict hierarchical model. In this way, our model acts as an {\em iterative refinement architecture} that reconsiders the input in each layer while being informed by the previous layer through initialisation.
This creates a hierarchy of refinement layers and each of them contributes to the NLI classification by max pooling the hidden states. In the following we refer to that architecture with the abbreviation HBMP. Max pooling is defined in the standard way of taking the highest value over each dimension of the hidden states and the final sentence embedding is the concatenation of those vectors coming from each BiLSTM layer. The overall architecture is illustrated in Figure~\ref{fig:architecture}.

To summarize the differences between our model and traditional stacked BiLSTM architectures we can list the following three main aspects:
\begin{enumerate}
    \item Each layer in our model is a separate BiLSTM initialized with the hidden and cell states of the previous layer.
    \item Each layer in our model receives the same word embeddings as its input.
    \item The final sentence representation is the concatenation of the max pooled output of each layer in the encoder network.
\end{enumerate}

In order to study the effect of our architecture we conduct a comparison of HBMP with the following alternative models:
\begin{enumerate}
    \item \textbf{BiLSTM-Ens:} Ensemble of three BiLSTMs with max pooling, all getting the same embeddings as the input.
    \item \textbf{BiLSTM-Ens-Train:} Ensemble of three BiLSTMs with max pooling, with the hidden and cell states of each BiLSTM being trainable parameters of the whole network.
    \item \textbf{BiLSTM-Ens-Tied:} Ensemble of three BiLSTMs with max pooling, where the weights of the BiLSTMs are tied.
    \item \textbf{BiLSTM-Stack:} A strictly hierarchical model with three BiLSTM layers where the second and third layer receive the output of the previous layer as their input.
\end{enumerate}

In the first model (BiLSTM-Ens) we contrast our architecture with a similar setup that does not transfer knowledge between layers but also combines information from three separate BiLSTM layers for the final classification. The second model (BiLSTM-Ens-Train) adds a trainable initialization to each layer to study the impact of the hierarchical initialization that we propose in our architecture. The third model (BiLSTM-Ens-Tied) connects the three layers by tying parameters to each other. Finally, the fourth model (BiLSTM-Stack) implements a standard hierarchical network with stacked layers that do not re-read the original input.

We apply the standard SNLI data for the comparison of these different architectures (see Section \ref{sec:setup} for more information about the SNLI benchmark). Table \ref{table:variants} lists the results of the experiment.

\begin{table}[h]
\begin{small}
\begin{center}
\begin{tabular}{l c c}
\hline \bf Model & \bf Accuracy & \bf Confidence Interval (95\%)*\\ \hline
600D HBMP (our model) & \bf 86.6 & [84.6\%, 88.7\%]\\
600D BiLSTM-Ens & 86.3 & [84.4\%, 88.3\%]\\
600D BiLSTM-Ens-Train & 86.3 & [84.3\%, 88.4\%]\\
600D BiLSTM-Ens-Tied & 86.1 & [83.8\%, 87.9\%]\\
600D BiLSTM-Stack & 86.3 & [84.2\%, 88.3\%]\\
\hline
\end{tabular}
\end{center}
\end{small}
\caption{\label{table:variants} SNLI test accuracies (\%) of different architectures. *Confidence intervals calculated over 1000 random samples of 1000 sentence pairs.}
\end{table}

The results show that HBMP performs better than each of the other models, which supports the use of our setup in favor of alternative architectures. Furthermore, we can see that the different components all contribute to the final score. Ensembling information from three {\em separate} BiLSTM layers (with independent parameters) improves the performance as we can see in the comparison between BiLSTM-Ens and BiLSTM-Ens-Tied. Trainable initialization does not seem to add to the model's capacity and indicates that the hierarchical initialization that we propose is indeed beneficial. Finally, feeding the same input embeddings to all BiLSTMs of HBMP leads to an improvement over the stacked model that does not re-read the input information.

Using these initial findings, we will now look at a more detailed analyses of the performance of HBMP on various datasets and tasks. But before, we first give some more details about the implementation of the model and the training procedures we use. Note, that the same specifications also apply to the experiments that we already discussed above.

\section{Training Details}

The architecture was implemented using PyTorch. We have published our code in GitHub: \url{https://github.com/Helsinki-NLP/HBMP}.

For all of our models we used a gradient descent optimization algorithm based on the Adam update rule \citep{KingmaB14Adam}, which is pre-implemented in PyTorch. We used a learning rate of 5e-4 for all our models. The learning rate was decreased by the factor of 0.2 after each epoch if the model did not improve. We used a batch size of 64. The models were evaluated with the development data after each epoch and training was stopped if the development loss increased for more than 3 epochs. The model with the highest development accuracy was selected for testing.

We use pre-trained GloVe word embeddings of size 300 dimensions (GloVe 840B 300D; \citealp{pennington2014glove}), which were fine-tuned during training. The sentence embeddings have hidden size of 600 for both direction (except for SentEval test, where we test models with 600D and 1200D per direction) and the 3-layer multi-layer perceptron (MLP) have the size of 600 dimensions. We use a dropout of 0.1 between the MLP layers (except just before the final layer). Our models were trained using one NVIDIA Tesla P100 GPU.


\section{Evaluation Benchmarks}
\label{sec:setup}

To further study the performance of HBMP, we
train our architecture with three common NLI datasets:
\begin{itemize}
\item the Stanford Natural Language Inference (SNLI) corpus,
\item the Multi-Genre Natural Language Inference (MultiNLI) corpus,
\item the Textual Entailment Dataset from Science Question Answering (SciTail).
\end{itemize}
Note that we treat them as separate tasks and do not mix any of the training, development and test data in our NLI experiments. We further perform additional linguistic error analyses using the {\em MultiNLI Annotation Dataset} and the {\em Breaking NLI} dataset. Finally, in order to test the ability of the model to learn general-purpose representations, we apply the downstream tasks that are bundled in the SentEval package for sentence embedding evaluation.
Note that we combine SNLI and MultiNLI data in those experiments in order to be compatible with related work.
Below we provide a few more details about each of the evaluation frameworks.

\paragraph{SNLI:}
The Stanford Natural Language Inference (SNLI) corpus \citep{snli} is a dataset of 570k human-written sentence pairs manually labeled with the gold labels entailment, contradiction, and neutral. The dataset is divided into training (550,152 pairs), development (10,000 pairs) and test sets (10,000 pairs). The source for the premise sentences in SNLI were image captions taken from the Flickr30k corpus \citep{Flickr:TACL229}.

\paragraph{MultiNLI:}
The Multi-Genre Natural Language Inference (MultiNLI) corpus \citep{multinli} is a broad-coverage corpus for natural language inference, consisting of 433k human-written sentence pairs labeled with entailment, contradiction and neutral. Unlike the SNLI corpus, which draws the premise sentence from image captions, MultiNLI consists of sentence pairs from ten distinct genres of both written and spoken English. The dataset is divided into training (392,702 pairs), development (20,000 pairs) and test sets (20,000 pairs).

Only five genres are included in the training set. The development and test sets have been divided into matched and mismatched, where the former includes only sentences from the same genres as the training data, and the latter includes sentences from the remaining genres not present in the training data.

In addition to the training, development and test sets, MultiNLI provides a smaller annotation dataset, which contains approximately 1000 sentence pairs annotated with linguistic properties of the sentences and is split between the matched and mismatched datasets.\footnote{The annotated dataset and description of the annotations are available at \url{http://www.nyu.edu/projects/bowman/
multinli/multinli_1.0_annotations.zip}} This dataset provides a simple way to assess what kind of sentence pairs an NLI system is able to predict correctly and where it makes errors. We use the annotation dataset to perform linguistic error analysis of our model and compare the results to results obtained with InferSent. For our experiment with the annotation dataset we use the annotations for the MultiNLI mismatched dataset.

\paragraph{SciTail:}
SciTail \citep{scitail} is an NLI dataset created from multiple-choice science exams consisting of 27k sentence pairs. Each question and the correct answer choice have been converted into an assertive statement to form the hypothesis. The dataset is divided into training (23,596 pairs), development (1,304 pairs) and test sets (2,126 pairs). Unlike the SNLI and MultiNLI datasets, SciTail uses only two labels: entailment and neutral.

\paragraph{Breaking NLI:}
Breaking NLI \citep{breakingNLI} is a test set (8,193 pairs) which is constructed by taking premises from the SNLI training set and constructing several hypotheses from them by changing at most one word within the premise. It was constructed to highlight how poorly current neural network models for NLI can handle lexical meaning.

\paragraph{SentEval:}
SentEval \citep{infersent,SentEval} is a library for evaluating the quality of sentence embeddings.\footnote{The SentEval test suite is available online at \url{https://github.com/facebookresearch/SentEval}.} It contains 17 downstream tasks as well as 10 probing tasks. The downstream datasets included in the tests were MR movie reviews, CR product reviews, SUBJ subjectivity status, MPQA opinion-polarity, SST binary sentiment analysis, TREC question-type classification, MRPC paraphrase detection, SICK-Relatedness (SICK-R) semantic textual similarity, SICK-Entailment (SICK-E) natural language inference and STS14 semantic textual similarity. The probing tasks evaluate how well the sentence encodings are able to capture the following linguistic properties: Length prediction, Word Content analysis, Tree depth prediction, Top Constituents prediction, Word order analysis, Verb tense prediction, Subject number prediction, Object number prediction, Semantic odd man out and Coordination Inversion.

For the SentEval tasks we trained our model on NLI data consisting of the concatenation of the SNLI and MultiNLI training sets consisting of 942,854 sentence pairs in total.  This allows us to compare our results to the InferSent results which were obtained using a model trained on the same data \citep{infersent}. \cite{infersent} have shown that including all the training data from SNLI and MultiNLI improves significantly the model performance on transfer learning tasks, compared to training the model only on SNLI data.

\section{Model Performance on the NLI task}

In this section, we discuss the performance of the proposed sentence-encoding approach in common natural language inference benchmarks. From the experiments, we can conclude that the model provides strong results on all of the three NLI datasets. It clearly outperforms the similar but non-hierarchical BiLSTM models reported in the literature and fares well in comparison to other state of the art architectures in the sentence encoding category. In particular, our results are close to the current state of the art on SNLI in this category and strong on both, the matched and mismatched test sets of MultiNLI. Finally, on SciTail, we achieve the new state of the art with an accuracy of 86.0\%.

Below, we provide additional details on our results for each of the benchmarks. We compare our model only with other state-of-the-art sentence encoding models and exclude cross-sentence attention models, except for SciTail where previous sentence encoding model-based results have not been published.

\subsection{SNLI}

For the SNLI dataset, our model provides the test accuracy of 86.6\% after 4 epochs of training. The comparison of our results with the previous state of the art and selected other sentence embedding based results are reported in Table \ref{table:SNLIresults}.

\begin{table}[h]
\begin{small}
\begin{center}
\begin{tabular}{l c}
\hline \bf Model & \bf Accuracy \\ \hline
BiLSTM Max Pool (InferSent)\textsuperscript{a} & 84.5\\
Distance-based Self-Attention\textsuperscript{b} & 86.3 \\
ReSA\textsuperscript{c} & 86.3 \\
600D BiLSTM with generalized pooling\textsuperscript{b} & 86.6 \\
600D Dynamic Self-Attention Model\textsuperscript{c} & 86.8 \\
2400D Multiple-Dynamic Self-Attention Model\textsuperscript{c} & \bf 87.4 \\
\hline
600D HBMP (our model) & 86.6\\
\hline
\end{tabular}
\end{center}
\end{small}
\caption{\label{table:SNLIresults} SNLI test accuracies (\%). Results marked with \textsuperscript{a} by \cite{infersent}, \textsuperscript{b} by \cite{Chen18} and \textsuperscript{c} by \cite{yoon2018dynattn}.}
\end{table}

\subsection{MultiNLI}

For the MultiNLI matched test set (MultiNLI-m) our model achieves a test accuracy of 73.7\% after 3 epochs of training, which is 0.8\% points lower than the state of the art 74.5\% by \citet{Nie:repeval}. For the mismatched test set (MultiNLI-mm) our model achieves a test accuracy of 73.0\% after 3 epochs of training, which is 0.6\% points lower than the state of the art 73.6\% by \citet{chen:repeval}.

\begin{table*}[t]
\begin{small}
\begin{center}
\begin{tabular}{l c c}
\hline
\bf  & \bf Accuracy  & \bf Accuracy  \\
\bf Model & \bf  (MultiNLI-m) & \bf (MultiNLI-mm) \\ \hline
CBOW\textsuperscript{a} & 66.2 & 64.6 \\
BiLSTM\textsuperscript{a} & 67.5 & 67.1 \\
BiLSTM + enh embed + max pooling\textsuperscript{b} & 70.7 & 70.8 \\
BiLSTM + Inner-attention\textsuperscript{c} & 72.1 & 72.1 \\
Deep Gated Attn. BiLSTM encoders\textsuperscript{d} & 73.5 & \bf 73.6 \\
Shortcut-Stacked BiLSTM\textsuperscript{e} & \bf 74.5 & 73.5 \\
\hline
600D HBMP & 73.7 & 73.0 \\
\hline
\end{tabular}
\end{center}
\end{small}
\caption{\label{table:MultiNLIresults} MultiNLI test accuracies (\%). Results marked with \textsuperscript{a} are baseline results by \cite{multinli}, \textsuperscript{b} by \cite{Vu:repeval}, \textsuperscript{c} by \cite{Balazs:repeval}, \textsuperscript{d} by \cite{chen:repeval} and \textsuperscript{e} by \cite{Nie:repeval}. Our results for the MultiNLI test sets were obtained by submitting the predictions to the respective Kaggle competitions.}
\end{table*}

A comparison of our results with the previous state of the art and selected other approaches are reported in Table \ref{table:MultiNLIresults}.

Although we did not achieve state of the art results for the MultiNLI dataset, we believe that a systematic study of different BiLSTM max pooling structures could reveal an architecture providing the needed improvement.

\subsection{SciTail}
\label{sec:ResultsSciTail}
On the SciTail dataset we compared our model also against non-sentence embedding-based models, as no results have been previously published which are based on independent sentence embeddings.
We obtain a score of 86.0\% after 4 epochs of training, which is +2.7\% points absolute improvement on the previous published state of the art by \cite{cafe}. Our model also outperforms InferSent which achieves an accuracy of 85.1\% in our experiments. The comparison of our results with the previous state of the art results are reported in Table \ref{table:SciTailresults}.

\begin{table}[h!]
\begin{small}
\begin{center}
\begin{tabular}{l c}
\hline \bf Model & \bf Accuracy \\ \hline
DecompAtt\textsuperscript{a} & 72.3\\
ESIM\textsuperscript{a} & 70.6 \\
Ngram\textsuperscript{a} & 70.6 \\
DGEM w/o edges\textsuperscript{a} & 70.8 \\
DGEM\textsuperscript{a} & 77.3 \\
CAFE\textsuperscript{b} & 83.3 \\
InferSent & 85.1 \\
\hline
600D HBMP & \bf 86.0 \\
\hline
\end{tabular}
\end{center}
\end{small}
\caption{\label{table:SciTailresults} SciTail test accuracies (\%).  Results marked with \textsuperscript{a} are baseline results reported by \cite{scitail} and \textsuperscript{b} by \cite{cafe}.}
\end{table}

The results achieved by our proposed model are significantly higher than the previously published results. It has been argued that the lexical similarity of the sentences in SciTail sentence pairs make it a particularly difficult dataset \citep{scitail}. If this is the case, we hypothesize that our model is indeed better at identifying entailment relations beyond focusing on the lexical similarity of the sentences.

\section{Error Analysis of NLI Predictions}

To better understand what kind of inferential relationships our model is able to identify, we conducted an error analysis for the three datasets. We report the results below.

Table \ref{table:LabelAccuracy} shows the accuracy of predictions per label (in terms of F-scores) for the HBMP model and compares them to the InferSent model. This analysis shows that our model leads to a significant improvement over the outcome of the non-hierarchical model from previous work in almost all categories on all the three benchmarks. The only exception is the entailment score on SciTail, which is slightly below the performance of InferSent.

\begin{table}[ht!]
\begin{tabular}{l|cc|cc|cc|cc}
\hline
 & \multicolumn{2}{c|}{SNLI} & \multicolumn{2}{c|}{MultiNLI-m} &\multicolumn{2}{c|}{MultiNLI-mm} &\multicolumn{2}{c}{SciTail}
\\
& \tiny\bf HBMP &\tiny\bf InferSent &\tiny\bf HBMP &\tiny\bf InferSent &\tiny\bf HBMP &\tiny\bf InferSent &\tiny\bf HBMP &\tiny\bf InferSent \\
\hline
entailment & \bf 88.5 & 86.8 & \bf 77.2 & 74.4 &\bf 77.9 & 74.9 & 81.0 & \bf 81.3\\
contradiction & \bf 89.1 & 86.2 & \bf 75.3 & 71.8 &\bf 75.6 & 71.5 & -& -\\
neutral  & \bf 81.9  & 80.9 &  \bf 68.2 & 67.1 & \bf 68.6 & 65.4 &\bf 88.9 & 88.1\\
\hline
\end{tabular}
\caption{\label{table:LabelAccuracy} Model performance by F-score, comparing HBMP to InferSent \citep{infersent} (our implementation).}
\end{table}

To see in more detail how our HBMP model is able to classify sentence pairs with different labels and what kind of errors it makes, we summarize error statistics as confusion matrices for the different datasets.  They highlight the HBMP model's strong performance across all the labels.

On the \textbf{SNLI dataset} our model clearly outperforms InferSent on all labels in terms of precision and recall. Table \ref{table:snliconf} contains the confusion matrices for that dataset comparing HBMP to InferSent. The precision on contradiction exceeds 90\% for our model and reaches high recall values for both, entailment and contradiction. The performance is lower for neutral and the confusion of that label with both, contradiction and entailment is higher. However, HBMP still outperforms InferSent by a similar margin as for the other two labels.

\begin{table}[t]
\newcommand\items{3}   
\arrayrulecolor{white} 
{\small
\noindent\begin{tabular}{rc*{\items}{|E}|c*{\items}{|E}|c|}
\multicolumn{1}{c}{} &\multicolumn{1}{c}{} &\multicolumn{\items}{c}{\bf Predicted - HBMP} &
\multicolumn{1}{c}{} &\multicolumn{\items}{c}{\bf Predicted - InferSent} \\ \hhline{~*\items{|-}|}
\multicolumn{1}{c}{} &
\multicolumn{1}{c}{} &
\multicolumn{1}{c}{entail} &
\multicolumn{1}{c}{contradict} &
\multicolumn{1}{c}{neutral} &
\multicolumn{1}{c}{\em recall} &
\multicolumn{1}{c}{entail} &
\multicolumn{1}{c}{contradict} &
\multicolumn{1}{c}{neutral} &
\multicolumn{1}{c}{\em recall}  \\ \hhline{~*\items{|-}|}
\multirow{\items}{*}{\rotatebox{90}{\bf Gold}}
&entail     & 3047 & 58   &  263  & \multicolumn{1}{c}{\bf 90.5\%} &
              2967 & 95   &  306  & \multicolumn{1}{c}{\bf 88.1\%} \\ \hhline{~*\items{|-}|}
&contradict &  117 & 2840 &  280  & \multicolumn{1}{c}{\bf 87.7\%} &
               154 & 2756 &  327  & \multicolumn{1}{c}{\bf 85.1\%} \\ \hhline{~*\items{|-}|}
&neutral    &  357 &  240 & 2622  & \multicolumn{1}{c}{\bf 81.5\%} &
               346 &  302 & 2571  & \multicolumn{1}{c}{\bf 79.9\%} \\ \hhline{~*\items{|-}|}
\hline
& \multicolumn{1}{c}{\em precision} &
\multicolumn{1}{c}{\bf 86.5\%} &
\multicolumn{1}{c}{\bf 90.5\%} &
\multicolumn{1}{c}{\bf 82.8\%} &
\multicolumn{1}{c}{} &
\multicolumn{1}{c}{\bf 85.6\%} &
\multicolumn{1}{c}{\bf 87.4\%} &
\multicolumn{1}{c}{\bf 80.2\%} \\ \hhline{~*\items{|-}|}
\end{tabular}
}
\caption{\label{table:snliconf}SNLI confusion matrices for HBMP and InferSent.}
\end{table}

Unlike for the SNLI and both of the MultiNLI datasets, on the \textbf{SciTail dataset} our model is most accurate on sentence pairs labeled neutral, having an F-score 88.9\% compared to pairs marked with entailment, where the F-score was 81.0\%. InferSent has slightly higher accuracy on entailment, whereas HBMP outperforms InferSent on neutral. Table \ref{table:scitailconf} contains the confusion matrices for the SciTail dataset comparing the HBMP to InferSent. This analysis reveals that our model mainly suffers in recall on entailment detection whereas it performs well for neutral with respect to recall. It is difficult to say what the reason might be for the mismatch between the two systems but the overall performance of our architecture suggests that it is superior to the InferSent model even though the balance between precision and recall on individual labels is different.

\begin{table}[t]
\newcommand\items{2}   
\arrayrulecolor{white} 
{\small
\noindent\begin{tabular}{rc*{\items}{|G}|c*{\items}{|G}|c|}
\multicolumn{1}{c}{} &\multicolumn{1}{c}{} &\multicolumn{\items}{c}{\bf HBMP} &
\multicolumn{1}{c}{} &\multicolumn{\items}{c}{\bf InferSent} \\ \hhline{~*\items{|-}|}
\multicolumn{1}{c}{} &
\multicolumn{1}{c}{} &
\multicolumn{1}{c}{entail} &
\multicolumn{1}{c}{neutral} &
\multicolumn{1}{c}{\em recall} &
\multicolumn{1}{c}{entail} &
\multicolumn{1}{c}{neutral} &
\multicolumn{1}{c}{\em recall}  \\ \hhline{~*\items{|-}|}
\multirow{\items}{*}{\rotatebox{90}{\bf Gold}}
&entail     & 632 & 210   & \multicolumn{1}{c}{\bf 75.0\%}
			& 673 & 169   & \multicolumn{1}{c}{\bf 79.9\%} \\\hhline{~*\items{|-}|}
&neutral    &  88 &  1196  & \multicolumn{1}{c}{\bf 93.1\%}
			&  140 &  1144  & \multicolumn{1}{c}{\bf 89.1\%}\\ \hhline{~*\items{|-}|}
\hline
& \multicolumn{1}{c}{\em precision} &
\multicolumn{1}{c}{\bf 88.0\%} &
\multicolumn{1}{c}{\bf 85.0\%} &
\multicolumn{1}{c}{} &
\multicolumn{1}{c}{\bf 82.8\%} &
\multicolumn{1}{c}{\bf 87.1\%} & \\ \hhline{~*\items{|-}|}
\end{tabular}
}
\arrayrulecolor{black}
\caption{\label{table:scitailconf}SciTail confusion matrices for HBMP and InferSent based on the development set.}
\end{table}

The error analysis of the \textbf{MultiNLI dataset} is not standard as it cannot be based on test data.  As the labeled test data is not openly available for MultiNLI, we analyzed the error statistics for this dataset based on the development data.

For the matched dataset (MultiNLI-m) our model had a development accuracy of 74.1\%. For MultiNLI-m our model has the best accuracy on sentence pairs labeled with entailment, having an F-score  of 77.2\%. The model is also almost as accurate in predicting contradictions, with an F-score of 75.3\%. Similar to SNLI, our model is less effective on sentence pairs labeled with neutral, having an F-score of 68.2\% but, again, the HBMP model outperforms the InferSent on all the labels.
Table \ref{table:multinlimconf} contains the confusion matrices for the MultiNLI matched dataset comparing the HBMP to InferSent. Our model improves upon InferSent in all values of precision and recall, in some cases by a wide margin.

\begin{table}[t]
\newcommand\items{3}   
\arrayrulecolor{white} 
{\small
\noindent\begin{tabular}{rc*{\items}{|F}|c*{\items}{|F}|c|}
\multicolumn{1}{c}{} &\multicolumn{1}{c}{} &\multicolumn{\items}{c}{\bf Predicted - HBMP} &
\multicolumn{1}{c}{} &\multicolumn{\items}{c}{\bf Predicted - InferSent} \\ \hhline{~*\items{|-}|}
\multicolumn{1}{c}{} &
\multicolumn{1}{c}{} &
\multicolumn{1}{c}{entail} &
\multicolumn{1}{c}{contradict} &
\multicolumn{1}{c}{neutral} &
\multicolumn{1}{c}{\em recall} &
\multicolumn{1}{c}{entail} &
\multicolumn{1}{c}{contradict} &
\multicolumn{1}{c}{neutral} &
\multicolumn{1}{c}{\em recall}  \\ \hhline{~*\items{|-}|}
\multirow{\items}{*}{\rotatebox{90}{\bf Gold}}
&entail     & 2781 & 196  &  486  & \multicolumn{1}{c}{\bf 80.3\%}
			& 2614 & 278  &  587  & \multicolumn{1}{c}{\bf 75.1\%}\\ \hhline{~*\items{|-}|}
&contradict &  372 & 2354 &   514  & \multicolumn{1}{c}{\bf 72.7\%}
			&  449 & 2241 &   523  & \multicolumn{1}{c}{\bf 69.7\%}\\ \hhline{~*\items{|-}|}
&neutral    &  528 &  443 & 2158  & \multicolumn{1}{c}{\bf 69.0\%}
			&  477 &  507 & 2139 & \multicolumn{1}{c}{\bf 68.5\%}\\ \hhline{~*\items{|-}|}
\hline
& \multicolumn{1}{c}{\em precision} &
\multicolumn{1}{c}{\bf 75.6\%} &
\multicolumn{1}{c}{\bf 78.7\%} &
\multicolumn{1}{c}{\bf 68.3\%} &
\multicolumn{1}{c}{} &
\multicolumn{1}{c}{\bf 73.8\%} &
\multicolumn{1}{c}{\bf 74.1\%} &
\multicolumn{1}{c}{\bf 65.8\%} &
\\ \hhline{~*\items{|-}|}
\end{tabular}
}
\arrayrulecolor{black}
\caption{\label{table:multinlimconf}MultiNLI-matched confusion matrices for HBMP and InferSent based on the development set.}
\end{table}

For the MultiNLI mismatched dataset (MultiNLI-mm) our model had a development accuracy of 73.7\%.
or MultiNLI-mm our model has very similar performance as with the MultiNLI-m dataset, having the best accuracy on sentence pars labeled with entailment, having an F-score of 77.9\%. The model is also almost as accurate in predicting contradictions, with an F-score of 75.6\%. Our model is less effective on sentence pairs labeled with neutral, having an F-score of 68.6\%. Table \ref{table:multinlimmconf} contains the confusion matrices for the MultiNLI Mismatched dataset comparing the HBMP to InferSent and the picture is similar to the result of the matched dataset. Substantial improvements can be seen again, in particular in the precision of contradiction detection.

\begin{table}[t]
\newcommand\items{3}   
\arrayrulecolor{white} 
{\small
\noindent\begin{tabular}{rc*{\items}{|F}|c*{\items}{|F}|c|}
\multicolumn{1}{c}{} &\multicolumn{1}{c}{} &\multicolumn{\items}{c}{\bf Predicted - HBMP} &
\multicolumn{1}{c}{} &\multicolumn{\items}{c}{\bf Predicted - InferSent} \\ \hhline{~*\items{|-}|}
\multicolumn{1}{c}{} &
\multicolumn{1}{c}{} &
\multicolumn{1}{c}{entail} &
\multicolumn{1}{c}{contradict} &
\multicolumn{1}{c}{neutral} &
\multicolumn{1}{c}{\em recall} &
\multicolumn{1}{c}{entail} &
\multicolumn{1}{c}{contradict} &
\multicolumn{1}{c}{neutral} &
\multicolumn{1}{c}{\em recall}  \\ \hhline{~*\items{|-}|}
\multirow{\items}{*}{\rotatebox{90}{\bf Gold}}
&entail     & 2841 & 163  &  459  & \multicolumn{1}{c}{\bf 82.0\%}
				& 2731 & 246  &  486  & \multicolumn{1}{c}{\bf 78.9\%}\\ \hhline{~*\items{|-}|}
&contradict  &  438 & 2279 &  523  & \multicolumn{1}{c}{\bf 70.3\%}
				&  491 & 2226 &  523  & \multicolumn{1}{c}{\bf 68.7\%}\\ \hhline{~*\items{|-}|}
&neutral        &  613 &  387 & 3111  & \multicolumn{1}{c}{\bf 68.0\%}
				&  611 &  510 & 2008  & \multicolumn{1}{c}{\bf 64.2\%}\\ \hhline{~*\items{|-}|}
\hline
& \multicolumn{1}{c}{\em precision} &
\multicolumn{1}{c}{\bf 73.0\%} &
\multicolumn{1}{c}{\bf 81.0\%} &
\multicolumn{1}{c}{\bf 68.4\%} &
\multicolumn{1}{c}{} &
\multicolumn{1}{c}{\bf 71.2\%} &
\multicolumn{1}{c}{\bf 74.6\%} &
\multicolumn{1}{c}{\bf 66.6\%} & \\ \hhline{~*\items{|-}|}
\end{tabular}
}
\arrayrulecolor{black}
\caption{\label{table:multinlimmconf}MultiNLI-mismatched confusion matrices for HBMP and InferSent.}
\end{table}

\section{Evaluation of Linguistic Abstractions}

The most interesting part of the sentence encoder approach to NLI is the ability of the system to learn generic sentence embeddings that capture abstractions, which can be useful for other downstream tasks as well. In order to understand the capabilities of our model we first look at the type of linguistic reasoning that the NLI system is able to learn using the MultiNLI annotation set and the Breaking NLI test set. Thereafter, we evaluate downstream tasks using the SentEval library to study the use of our NLI-based sentence embeddings in transfer learning.

\subsection{Linguistic Error Analysis of NLI Classifications}

The MultiNLI annotation set makes it possible to conduct a detailed analysis of different linguistic phenomena when predicting inferential relationships. We use this to compare our model to InferSent with respect to the type of linguistic properties that are present in the given sentence pairs.
Table \ref{table:MismatchedLinguisticErrors} contains the comparison for the MultiNLI-mm dataset. The analysis shows that our HBMP model outperforms InferSent with antonyms, coreference links, modality, negation, paraphrases and tense differences. It also produces improved scores for most of the other categories in entailment detection. InferSent gains especially with conditionals in contradiction and in the word overlap catehory for entailments. This seems to suggest that InferSent relies a lot on matching words to find entailment and specific constructions indicating contradictions. HBMP does not seem to use word overlap as an indication for entailment that much and is better on detecting neutral sentences in this category. This outcome may indicate that our model works with stronger lexical abstractions than InferSent.
However, due to the small number of examples per annotation category and small differences in the scores in general, it is hard to draw reliable conclusions from this experiment.

\begin{table*}[h!]
\begin{center}
\scalebox{.9}{\begin{tabular}{l|cc|cc|cc}%
\hline
& \multicolumn{2}{c|}{Entailment} & \multicolumn{2}{c|}{Contradiction}  & \multicolumn{2}{c}{Neutral} \\
& \bf HBMP & \bf InferSent & \bf HBMP & \bf InferSent & \bf HBMP & \bf InferSent\\
\hline
active/passive (10) & 100.0 & 100.0 & 100.0 & 100.0 & - & -\\
anto (16) & - & - & \bf 76.9 & 69.2 &\bf 85.7 & 71.4\\
belief (44) & \bf 88.2 & 82.4 & \bf 66.7 & 61.1 & 73.9 & \bf 78.3\\
conditional (16) & 81.8 & 81.8 & 37.5 & \bf 62.5 & 57.1 & \bf 71.4\\
coref (22) & 75.0 & 75.0 & \bf 71.4 & 64.3 & 81.8 & 81.8\\
long sentence (77) &\bf  80.6 & 77.4 & 58.3 &\bf  61.1 &\bf  73.8 & 71.4\\
modal (98) &\bf  80.9 & 78.7 & 68.6 & 68.6 &\bf  81.8 & 70.5\\
negation (78) &\bf 76.0 & 64.0 & \bf 81.8 & 76.4 &\bf  58.3 & 45.8\\
paraphrase (33) &\bf 89.2 & 86.5 & - & - &  & -\\
quantifier (104) & \bf 75.0 & 72.5 & 73.1 & 73.1 & 75.0 & \bf 77.1 \\
quantity/time (15) & 33.3 & \bf 50.0 & 41.7 & 41.7 & 33.3 & 33.3\\
tense difference (14) & \bf 100.0 & 75.0 & 0.0 & 0.0 & \bf 83.3 & 75.0\\
word overlap (26) & 90.5 & \bf 95.2 & 41.7 & 41.7 & \bf 50.0 & 0.0\\
\hline
Total & 80.6 & \bf 83.0 & 66.3 & \bf 68.5 & \bf 73.8 & 72.7\\
\hline
\end{tabular}}
\end{center}
\caption{\label{table:MismatchedLinguisticErrors} MultiNLI-mm linguistic error analysis (accuracy \%), comparing our HBMP results to the InferSent \cite{infersent} results (our implementation). Number of sentence pairs with the linguistic label in brackets after the label name.}
\end{table*}

\subsection{Tests with the Breaking NLI dataset}

In the second experiment we conducted testing of the proposed sentence embedding architecture using the \textbf{Breaking NLI test set} recently published by \cite{breakingNLI}. The test set is designed to highlight the lack of lexical reasoning capability of NLI systems.

For the Breaking NLI experiment, we trained our HBMP model and the InferSent model using the SNLI training data. We compare our results with the results published by \cite{breakingNLI} and to results obtained with InferSent sentence encoder (our implementation).

The results show that our HBMP model outperforms the InferSent model in 7 out of 14 categories, receiving an overall score of 65.1\% (InferSent: 65.6\%). Our model is especially strong with handling antonyms, which shows a good level of semantic abstraction on the lexical level.
InferSent fares well in narrow categories like drinks, instruments and planets, which may indicate a problem of overfitting to prominent examples in the training data. The strong result on the synonyms class may also come from a significant representation of related examples in training. However, more detailed investigations are necessary to verify this hypothesis.

Our model also compares well against the other models, outperforming Decomposable Attention model (51.90\%) \citep{ParikhT0U16} and Residual Encoders (62.20\%) \citep{Nie:repeval} in the overall score. As these models are not based purely on sentence embeddings, the obtained result highlights that sentence embedding approaches can be competitive when handling inferences requiring lexical information. The results of the comparison are summarized in Table \ref{table:BreakingNLI}.

\begin{table}[ht!]
\begin{center}
\scalebox{.9}{\begin{tabular}{lcccc|cc}%
\hline

& \bf Decomp.  & \bf   & \bf WordNet & & \bf Infer- & \bf 600D \\
\bf Category & \bf Attn\textsuperscript{*} & \bf ESIM\textsuperscript{*} & \bf Baseline\textsuperscript{*} & \bf KIM\textsuperscript{*} & \bf Sent & \bf HBMP \\
\hline
antonyms & 41.6 & 70.4 & 95.5 & 86.5 & 51.6 & \bf 54.7 \\
antonyms(wordnet) & 55.1 & 74.6  & 94.5 & 78.8 & 63.7 & \bf 69.1 \\
cardinals & 53.5 & 75.5 & 98.6 & 93.4 & 49.4 & \bf 58.8 \\
colors & 85.0 & 96.1 & 98.7 & 98.3 & \bf 90.6 & 90.4 \\
countries & 15.2 & 25.4 & 100.0 & 70.8 & 77.2 & \bf 81.2 \\
drinks & 52.9 & 63.7 & 94.8 & 96.6 & \bf 85.1 & 81.3 \\
instruments & 96.9 & 90.8 & 67.7 & 96.9 & \bf 98.5 & 96.9 \\
materials & 65.2 & 89.7 & 75.3 & 98.7 & 81.6 & \bf 82.6 \\
nationalities & 37.5 & 35.9 & 78.5 & 73.5 & 47.3 & \bf 49.8 \\
ordinals & 2.1 & 21.0 & 40.7 & 56.6 & \bf 7.4 & 4.5 \\
planets & 31.7 & 3.3 & 100.0 & 5.0 & \bf 75.0 & 45.0 \\
rooms & 59.2 & 69.4 & 89.9 & 77.6 & \bf 76.3 & 72.1 \\
synonyms & 97.5 & 99.7 & 70.5 & 92.1 & \bf 99.6 & 84.5 \\
vegetables & 43.1 & 31.2 & 86.2 & 79.8 & 39.5 & \bf 40.4 \\
\hline
Total & 51.9 & 65.6 & 85.8 & 83.5 & \bf 65.6 & 65.1 \\
\hline
\end{tabular}}
\end{center}
\caption{\label{table:BreakingNLI}Breaking NLI scores (accuracy \%). Results marked with \textsuperscript{*} as reported by \cite{breakingNLI}. InferSent results obtained with our implementation using the training set-up described in \citep{infersent}. Scores highlighted with bold are top scores when comparing the InferSent and our HBMP model.}
\end{table}

\subsection{Transfer Learning}
\label{sec:Transfer}

In this section, we focus on transfer learning experiments that apply sentence embeddings trained on NLI to other downstream tasks.
In order to better understand how well the sentence encoding model generalizes to different tasks, we conducted various tests implemented in the SentEval sentence embedding evaluation library \citep{infersent} and compared our results to the results published for InferSent and SkipThought \citep{KirosZSZTUF15}.

\begin{table*}[ht!]
\begin{center}
\scalebox{.9}{\begin{tabular}{l|cc|cc}
\hline
\bf Task & \bf InferSent & \bf SkipThought & \bf 600D HBMP & \bf 1200D HBMP\\
\hline
\bf MR & 81.1 & 79.4 & 81.5 & \bf 81.7 \\
\bf CR & 86.3 & 83.1 & 86.4 & \bf 87.0 \\
\bf SUBJ & 92.4 & \bf 93.7 & 92.7 & \bf 93.7 \\
\bf MPQA & 90.2 & 89.3 & 89.8 & \bf 90.3 \\
\bf SST & \bf 84.6 & 82.9 & 83.6 & 84.0 \\
\bf TREC  & 88.2   & 88.4  & 86.4   & \bf 88.8 \\
\bf MRPC & 76.2/83.1 & - & 74.6/82.0 & \bf 76.7/83.4 \\
\bf SICK-R & \bf 0.884 & 0.858 & 0.876 & 0.876 \\
\bf SICK-E & \bf 86.3 & 79.5 & 85.3 & 84.7 \\
\bf STS14 & .70/.67 & .44/.45  & .70/.66 & \bf  .71/.68\\
\hline
\end{tabular}}
\end{center}
\caption{\label{table:TransferLearning} Transfer learning test results for the HBMP model on a number of SentEval downstream sentence embedding evaluation tasks. InferSent and SkipThought results as reported by \cite{infersent}. To remain consistent with other work using SentEval, we report the accuracies as they are provided by the SentEval library.}
\end{table*}

We used the SentEval library with the default settings recommended on their website, with a logistic regression classifier, Adam optimizer with learning rate of 0.001, batch size of 64 and epoch size of 4. Table \ref{table:TransferLearning} lists the transfer learning results for our models with 600D and 1200D hidden dimensionality and compares it to the InferSent and SkipThought scores reported by \cite{infersent}. Our 1200D model outperforms the InferSent model on 7 out of 10 tasks. The model achieves higher score on 8 out of 9 tasks reported for SkipThought, having equal score on the SUBJ dataset. No MRPC results have been reported for SkipThought.

\begin{table*}[ht!]
\begin{center}
\scalebox{.9}{
\bgroup
\setlength\tabcolsep{4pt}
\begin{tabular}{l|c|cc}
\hline
\bf Task & \bf InferSent & \bf 600D HBMP & \bf 1200D HBMP\\
\hline
\bf SentLen & 71.7 & \bf 75.9 & 75.0\\
\bf WC & \bf 87.3 & 84.1 & 85.3\\
\bf TreeDepth & 41.6 & 42.9 & \bf 43.8\\
\bf TopConst & 70.5 & 76.6 & \bf 77.2\\
\bf BShift & 65.1 & 64.3 & \bf 65.6\\
\bf Tense & 86.7 & 86.2 & \bf 88.0\\
\bf SubjNum & 80.7 & 83.7 & \bf 87.0\\
\bf ObjNum & 80.3 & 79.3 & \bf 81.8\\
\bf SOMO & \bf 62.1 & 58.9 & 59.0\\
\bf CoordInv & 66.8 & 68.5 & \bf 70.8\\
\hline
\end{tabular}
\egroup
}
\end{center}
\caption{\label{table:Probing} SentEval probing task results (accuracy \%). InferSent results are BiLSTM Max (NLI) results as reported by \cite{probing}.}
\end{table*}

To study in more detail the linguistic properties of our proposed model, we also ran the recently published SentEval probing tasks \citep{probing}. Our 1200D model outperforms the InferSent model in 8 out of 10 probing tasks. The results are listed in Table \ref{table:Probing}.

Looking at both the downstream and the probing tasks we can observe strong results of our model compared to the InferSent model that already demonstrated good general abstractions on the sentence level according to the original publication by \cite{infersent}. Hence, HBMP does not only provide competitive NLI scores but also produces improved sentence embeddings that are useful for other tasks.

\section{Conclusion}

In this paper we have introduced an iterative refinement architecture (HBMP) based on BiLSTM layers with max pooling that achieves a new state of the art for SciTail and strong results in the SNLI and MultiNLI sentence-encoding category. We carefully analyzed the performance of our model with respect to the label categories and the errors it produces in the various NLI benchmarks. We demonstrate that our model outperforms InferSent in nearly all cases with substantially reduced confusion between classes of inferential relationships.
The linguistic analysis on MultiNLI also reveals that our approach is robust across the various categories and outperforms InferSent on, for example, antonyms and negations that require a good level of semantic abstraction.

Furthermore, we tested our model using the SentEval sentence embedding evaluation library, showing that it achieves great generalization capability.  The model outperforms InferSent on 7 out of 10 downstream and 8 out of 10 probing tasks, and SkipThought on 8 out of 9 downstream tasks.   Overall, our model performs well across all the conducted experiments, which highlights its applicability for various NLP tasks and further demonstrates the general abstractions that it is able to pick up from the NLI training data.

Although the neural network approaches to NLI have been hugely successful, there has also been a number of concerns raised about the quality of current NLI datasets. \citet{gururangan:2018} and \citet{poliak2018} show that datasets like SNLI and MultiNLI contain annotation artifacts which help neural network models in classification, allowing decisions only based on the hypothesis sentences as their input. On a theoretical and methodological level, there is an on-going discussion on the nature of various NLI datasets, as well as the definition of what counts as NLI and what does not. For example, \citet{chatzikyriakidis2017} present an overview of the most standard datasets for NLI and show that the definitions of inference in each of them are actually quite different. \citet{talman2018testing} further highlight this by testing different state-of-the-art neural network models by training them on one dataset and then testing on another, leading to a significant drop in performance for all models.

In addition to the concerns related to the quality of NLI datasets, the success of the proposed architecture raises a number of other interesting questions. First of all, it would be important to understand what kind of semantic information the different layers are able to capture and how they differ from each other. Secondly, we would like to ask whether other architecture configurations could lead to even stronger results in NLI and other downstream tasks. A third question is concerned with other languages and cross-lingual settings. Does the result carry over to multilingual setups and applications? The final question is whether NLI-based sentence embeddings could successfully be combined with other supervised and also unsupervised ways of learning sentence-level representations. We will look at all those questions in our future work.

\section*{Acknowledgments}
The work in this paper was supported by the Academy of Finland through project 314062 from the ICT 2023 call on Computation, Machine Learning and Artificial Intelligence, and through projects 270354/273457/313478.

 \begin{wrapfigure}{r}{0.3\hsize}
 \flushleft
   \vspace{-20pt} \hspace{10pt}
   \includegraphics[width=.7\hsize]{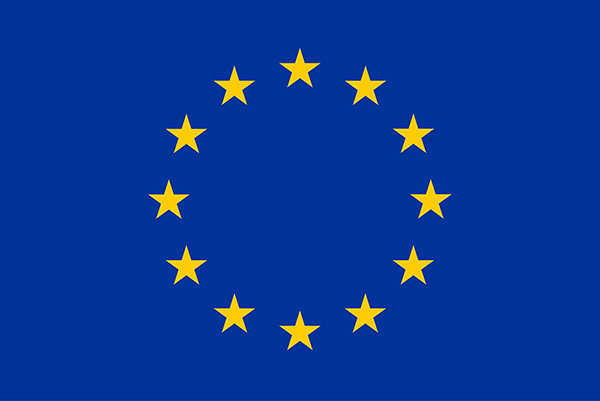}
   \vspace{-10pt} \hspace{-10pt}
\end{wrapfigure}

This project has also received funding from the European Research Council (ERC) under the European Union’s Horizon 2020 research and innovation programme (grant agreement No 771113).

We would also like to acknowledge NVIDIA and their GPU grant.

\bibliographystyle{chicago-nle}
\bibliography{HBMP}

\begin{thebibliography}{}

\bibitem[\protect\citeauthoryear{Balazs, Marrese-Taylor, Loyola, and
  Matsuo}{Balazs et~al.}{2017}]{Balazs:repeval}
Balazs, J., E.~Marrese-Taylor, P.~Loyola, and Y.~Matsuo 2017.
\newblock Refining raw sentence representations for textual entailment
  recognition via attention.
\newblock In {\em Workshop on Evaluating Vector Space Representations for NLP}.
  ACL.

\bibitem[\protect\citeauthoryear{Bowman, Angeli, Potts, and Manning}{Bowman
  et~al.}{2015}]{snli}
Bowman, S.~R., G.~Angeli, C.~Potts, and C.~D. Manning 2015.
\newblock A large annotated corpus for learning natural language inference.
\newblock In {\em EMNLP}.

\bibitem[\protect\citeauthoryear{Bowman, Gauthier, Rastogi, Gupta, Manning, and
  Potts}{Bowman et~al.}{2016}]{BowmanGRGMP16}
Bowman, S.~R., J.~Gauthier, A.~Rastogi, R.~Gupta, C.~D. Manning, and C.~Potts
  2016.
\newblock A fast unified model for parsing and sentence understanding.
\newblock In {\em ACL}.

\bibitem[\protect\citeauthoryear{Chatzikyriakidis, Cooper, Dobnik, and
  Larsson}{Chatzikyriakidis et~al.}{2017}]{chatzikyriakidis2017}
Chatzikyriakidis, S., R.~Cooper, S.~Dobnik, and S.~Larsson 2017.
\newblock An overview of natural language inference data collection: The way
  forward?
\newblock In {\em Computing Natural Language Inference Workshop}.

\bibitem[\protect\citeauthoryear{{Chen}, {Ling}, and {Zhu}}{{Chen}
  et~al.}{2018}]{Chen18}
{Chen}, Q., Z.-H. {Ling}, and X.~{Zhu} 2018.
\newblock {Enhancing Sentence Embedding with Generalized Pooling}.
\newblock In {\em COLING}.

\bibitem[\protect\citeauthoryear{Chen, Zhu, Ling, Wei, Jiang, and Inkpen}{Chen
  et~al.}{2017a}]{Chen}
Chen, Q., X.~Zhu, Z.-H. Ling, S.~Wei, H.~Jiang, and D.~Inkpen 2017a.
\newblock Enhanced lstm for natural language inference.
\newblock In {\em ACL}.

\bibitem[\protect\citeauthoryear{Chen, Zhu, Ling, Wei, Jiang, and Inkpen}{Chen
  et~al.}{2017b}]{chen:repeval}
Chen, Q., X.~Zhu, Z.-H. Ling, S.~Wei, H.~Jiang, and D.~Inkpen 2017b.
\newblock Recurrent neural network-based sentence encoder with gated attention
  for natural language inference.
\newblock In {\em Workshop on Evaluating Vector Space Representations for NLP}.
  ACL.

\bibitem[\protect\citeauthoryear{Conneau and Kiela}{Conneau and
  Kiela}{2018}]{SentEval}
Conneau, A. and D.~Kiela 2018, May 7--12.
\newblock Sent{E}val: An evaluation toolkit for universal sentence
  representations.
\newblock In N.~Calzolari (Ed.), {\em LREC 2018, Eleventh International
  Conference on Language Resources and Evaluation}, Phoenix Seagaia Conference
  Center, Miyazaki, Japan, pp.\  1699--1704.

\bibitem[\protect\citeauthoryear{Conneau, Kiela, Schwenk, Barrault, and
  Bordes}{Conneau et~al.}{2017}]{infersent}
Conneau, A., D.~Kiela, H.~Schwenk, L.~Barrault, and A.~Bordes 2017.
\newblock Supervised learning of universal sentence representations from
  natural language inference data.
\newblock In {\em EMNLP}.

\bibitem[\protect\citeauthoryear{Conneau, Kruszewski, Lample, Barrault, and
  Baroni}{Conneau et~al.}{2018}]{probing}
Conneau, A., G.~Kruszewski, G.~Lample, L.~Barrault, and M.~Baroni 2018.
\newblock {What you can cram into a single vector: Probing sentence embeddings
  for linguistic properties}.
\newblock In {\em ACL}.

\bibitem[\protect\citeauthoryear{Glockner, Shwartz, and Goldberg}{Glockner
  et~al.}{2018}]{breakingNLI}
Glockner, M., V.~Shwartz, and Y.~Goldberg 2018.
\newblock Breaking nli systems with sentences that require simple lexical
  inferences.
\newblock In {\em ACL}.

\bibitem[\protect\citeauthoryear{Gururangan, Swayamdipta, Levy, Schwartz,
  Bowman, and Smith}{Gururangan et~al.}{2018}]{gururangan:2018}
Gururangan, S., S.~Swayamdipta, O.~Levy, R.~Schwartz, S.~Bowman, and N.~A.
  Smith 2018.
\newblock Annotation artifacts in natural language inference data.
\newblock In {\em NAACL}. ACL.

\bibitem[\protect\citeauthoryear{Hill, Cho, and Korhonen}{Hill
  et~al.}{2016}]{HillCK16}
Hill, F., K.~Cho, and A.~Korhonen 2016.
\newblock Learning distributed representations of sentences from unlabelled
  data.
\newblock In {\em NAACL}.

\bibitem[\protect\citeauthoryear{Khot, Sabharwal, and Clark}{Khot
  et~al.}{2018}]{scitail}
Khot, T., A.~Sabharwal, and P.~Clark 2018.
\newblock Scitail: A textual entailment dataset from science question
  answering.
\newblock In {\em AAAI}.

\bibitem[\protect\citeauthoryear{Kingma and Ba}{Kingma and
  Ba}{2015}]{KingmaB14Adam}
Kingma, D.~P. and J.~Ba 2015.
\newblock Adam: {A} method for stochastic optimization.
\newblock In {\em ICLR}.

\bibitem[\protect\citeauthoryear{Kiros, Zhu, Salakhutdinov, Zemel, Urtasun,
  Torralba, and Fidler}{Kiros et~al.}{2015}]{KirosZSZTUF15}
Kiros, R., Y.~Zhu, R.~Salakhutdinov, R.~S. Zemel, R.~Urtasun, A.~Torralba, and
  S.~Fidler 2015.
\newblock Skip-thought vectors.
\newblock In {\em NeurIPS}.

\bibitem[\protect\citeauthoryear{Maas, Hannun, and Ng}{Maas
  et~al.}{2013}]{Maas2013RectifierNI}
Maas, A.~L., A.~Y. Hannun, and A.~Y. Ng 2013.
\newblock Rectifier nonlinearities improve neural network acoustic models.
\newblock In {\em Intl. Conf. on Machine Learning}.

\bibitem[\protect\citeauthoryear{Mikolov, Sutskever, Chen, Corrado, and
  Dean}{Mikolov et~al.}{2013}]{Mikolov:2013}
Mikolov, T., I.~Sutskever, K.~Chen, G.~Corrado, and J.~Dean 2013.
\newblock Distributed representations of words and phrases and their
  compositionality.
\newblock In {\em NeurIPS}, USA.

\bibitem[\protect\citeauthoryear{Mou, Men, Li, Xu, Zhang, Yan, and Jin}{Mou
  et~al.}{2016}]{MouConv}
Mou, L., R.~Men, G.~Li, Y.~Xu, L.~Zhang, R.~Yan, and Z.~Jin 2016.
\newblock Natural language inference by tree-based convolution and heuristic
  matching.
\newblock In {\em ACL}.

\bibitem[\protect\citeauthoryear{Nie and Bansal}{Nie and
  Bansal}{2017}]{Nie:repeval}
Nie, Y. and M.~Bansal 2017.
\newblock Shortcut-stacked sentence encoders for multi-domain inference.
\newblock In {\em Workshop on Evaluating Vector Space Representations for NLP}.
  ACL.

\bibitem[\protect\citeauthoryear{Parikh, T{\"{a}}ckstr{\"{o}}m, Das, and
  Uszkoreit}{Parikh et~al.}{2016}]{ParikhT0U16}
Parikh, A.~P., O.~T{\"{a}}ckstr{\"{o}}m, D.~Das, and J.~Uszkoreit 2016.
\newblock A decomposable attention model for natural language inference.
\newblock In {\em EMNLP}.

\bibitem[\protect\citeauthoryear{Pennington, Socher, and Manning}{Pennington
  et~al.}{2014}]{pennington2014glove}
Pennington, J., R.~Socher, and C.~D. Manning 2014.
\newblock Glove: Global vectors for word representation.
\newblock In {\em EMNLP}.

\bibitem[\protect\citeauthoryear{Poliak, Naradowsky, Haldar, Rudinger, and
  Van~Durme}{Poliak et~al.}{2018}]{poliak2018}
Poliak, A., J.~Naradowsky, A.~Haldar, R.~Rudinger, and B.~Van~Durme 2018.
\newblock Hypothesis only baselines in natural language inference.
\newblock In {\em Joint Conference on Lexical and Computational Semantics}.
  ACL.

\bibitem[\protect\citeauthoryear{Talman and Chatzikyriakidis}{Talman and
  Chatzikyriakidis}{2019}]{talman2018testing}
Talman, A. and S.~Chatzikyriakidis 2019.
\newblock Testing the generalization power of neural network models across nli
  benchmarks.
\newblock In {\em Proceedings of the 2019 {ACL} Workshop {B}lackbox{NLP}:
  Analyzing and Interpreting Neural Networks for {NLP}}.

\bibitem[\protect\citeauthoryear{Tay, Tuan, and Hui}{Tay et~al.}{2018}]{cafe}
Tay, Y., L.~A. Tuan, and S.~C. Hui 2018.
\newblock Compare, compress and propagate: Enhancing neural architectures with
  alignment factorization for natural language inference.
\newblock In {\em EMNLP}.

\bibitem[\protect\citeauthoryear{Vendrov, Kiros, Fidler, and Urtasun}{Vendrov
  et~al.}{2016}]{VendrovKFU15}
Vendrov, I., R.~Kiros, S.~Fidler, and R.~Urtasun 2016.
\newblock Order-embeddings of images and language.

\bibitem[\protect\citeauthoryear{Vu}{Vu}{2017}]{Vu:repeval}
Vu, H. 2017.
\newblock Lct-malta's submission to repeval 2017 shared task.
\newblock In {\em Workshop on Evaluating Vector Space Representations for NLP}.
  ACL.

\bibitem[\protect\citeauthoryear{Williams, Nangia, and Bowman}{Williams
  et~al.}{2018}]{multinli}
Williams, A., N.~Nangia, and S.~R. Bowman 2018.
\newblock A broad-coverage challenge corpus for sentence understanding through
  inference.
\newblock In {\em NAACL}.

\bibitem[\protect\citeauthoryear{Yoon, Lee, and Lee}{Yoon
  et~al.}{2018}]{yoon2018dynattn}
Yoon, D., D.~Lee, and S.~Lee 2018.
\newblock {Dynamic Self-Attention : Computing Attention over Words Dynamically
  for Sentence Embedding}.
\newblock {\em arXiv:1808.07383\/}.

\bibitem[\protect\citeauthoryear{Young, Lai, Hodosh, and Hockenmaier}{Young
  et~al.}{2014}]{Flickr:TACL229}
Young, P., A.~Lai, M.~Hodosh, and J.~Hockenmaier 2014.
\newblock From image descriptions to visual denotations: New similarity metrics
  for semantic inference over event descriptions.
\newblock {\em TACL\/}~{\em 2}.

\end{thebibliography}

\label{lastpage}

\end{document}